\documentclass[journal]{IEEEtran}

\usepackage{xcolor,soul,framed} 

\colorlet{shadecolor}{yellow}
\usepackage[pdftex]{graphicx}
\graphicspath{{../pdf/}{../jpeg/}}
\DeclareGraphicsExtensions{.pdf,.jpeg,.png}

\usepackage[cmex10]{amsmath}
\usepackage{array}
\usepackage{mdwmath}
\usepackage{mdwtab}
\usepackage{eqparbox}
\usepackage{url}
\usepackage{amssymb}
\usepackage{graphicx}
\usepackage{booktabs}
\usepackage{multirow}
\usepackage{algorithm}
\usepackage{algpseudocode}

\begin{document}
    \title{\textit{DeepForgeSeal:} Latent Space-Driven Semi-Fragile Watermarking for Deepfake Detection Using Adversarial Reinforcement Learning}

\author{Tharindu~Fernando,~\IEEEmembership{Member,~IEEE,}       Clinton~Fookes,~\IEEEmembership{Senior Member,~IEEE,}
     ~and~
        Sridha~Sridharan,~\IEEEmembership{Life Senior Member,~IEEE.}

\IEEEcompsocitemizethanks{\IEEEcompsocthanksitem T. Fernando, C. Fookes, and S. Sridharan are with The Signal Processing, Artificial Intelligence and Vision Technologies (SAIVT), Queensland University of Technology, Australia.\protect }}

\markboth{}{Fernando \MakeLowercase{\textit{et al.}}: working title}

\maketitle

\begin{abstract}
Rapid advances in generative AI have led to increasingly realistic deepfakes, posing growing challenges for law enforcement and public trust. Existing passive deepfake detectors struggle to keep pace, largely due to their dependence on specific forgery artifacts, which limits their ability to generalize to new deepfake types. Proactive deepfake detection using watermarks has emerged to address the challenge of identifying high-quality synthetic media. However, these methods often struggle to balance robustness against benign distortions with sensitivity to malicious tampering. This paper introduces a novel deep learning framework that harnesses high-dimensional latent space representations and the Adversarial Reinforcement Learning (ARL) paradigm to develop a robust and adaptive watermarking approach. Specifically, we develop a learnable watermark embedder that operates in the latent space, capturing high-level image semantics, while offering precise control over message encoding and extraction. The ARL paradigm empowers the learnable watermarking module to pursue an optimal balance between robustness and fragility. This is achieved through interaction with a dynamic curriculum of benign and malicious image manipulations simulated by an adversarial attacker agent. Comprehensive evaluations on the CelebA and CelebA-HQ benchmarks reveal that our method consistently outperforms state-of-the-art approaches, achieving improvements of over 4.5\% on CelebA and more than 5.3\% on CelebA-HQ under challenging manipulation scenarios.
\end{abstract}


\IEEEpeerreviewmaketitle

\section{Introduction}
\label{sec:intro}
Generative AI has rapidly transformed the way we create and edit multimedia, making it faster and easier than ever to produce high-quality images, audio, and video \cite{qiao2024fully}. While these innovations unlock exciting possibilities, they also introduce serious risks. Among the most concerning threats are image and video deepfakes, highly realistic, AI-generated images and videos that mimic real individuals. These deepfakes pose significant challenges, not only for their role in spreading misinformation and disinformation at an alarming pace but also for their implications on privacy, consent, security, and societal trust \cite{deepfakewashingtonpost,
deepfakebrookings1,
deepfakeNSA,
fernando2025face,
mustak2023deepfakes}. 

\begin{figure}[!t]
    \centering
    \includegraphics[width=\linewidth]{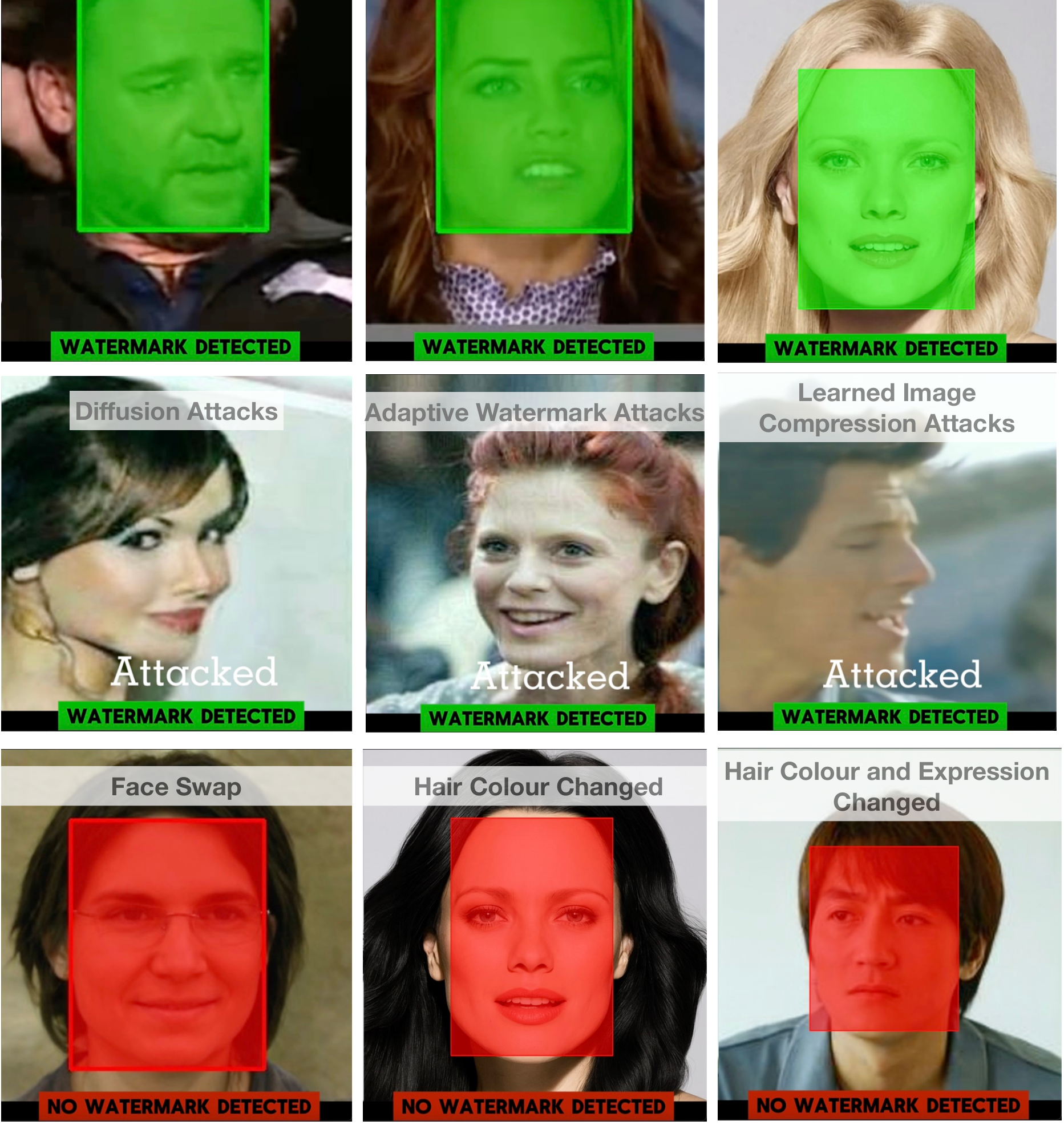}
    \caption{Watermarking and watermark retrieval performance of proposed \textit{DeepForgeSeal} framework. The visualisations include low-resolution and high-resolution bonafide images (row 1), numerous learned attacks against watermarking (row 2), and watermarked images that have undergone semantic-altering manipulations (row 3).}
    \label{fig:hero_figure}
\end{figure}

To combat the rapid spread of deepfakes, watermarking \cite{zhang2025facetracer, qu2025id} has emerged as a technique that allows users to validate the authenticity and integrity of the multimedia they consume. A robust watermarking technique should be resilient to distortions such as JPEG compression, cropping, and colour changes to keep the embedded watermark intact. While being resilient to such incidental distortions, a good watermarking technique should demonstrate fragility towards content-altering modifications, such as face-swapping-type malicious edits that alter the semantic-level meaning of the image. Recently, deep learning techniques \cite{zhao2023proactive, wang2025fractalforensics, wu2023sepmark, zhang2025omniguard, zhang2024editguard} have made significant progress towards achieving this paradoxical combination of resilience and fragility. For instance, SepMark \cite{wu2023sepmark} introduced an encoder–decoder architecture with a unified encoder for watermark embedding. It employs two distinct decoders to recover the watermark under different robustness constraints, enabling selective resilience to varying levels of adversarial distortion. However, most existing deep watermarking techniques for deepfake detection rely on a fixed, pre-defined watermark location for extraction. This design choice makes them vulnerable, as perpetrators can easily detect and tamper with the watermark once its position is known. In addition, these methods often lack robustness against benign image manipulations such as brightness adjustments or heavy compression, leading to a high rate of false positives in deepfake detection. Furthermore, current supervised and adversarial learning approaches provide only limited fidelity in addressing the resilience–fragility paradox.

Most recently, OmniGuard \cite{zhang2025omniguard} introduced a more flexible approach for image manipulation detection and localisation by incorporating a degradation-aware tamper extraction network designed for blind, robust copyright protection and tamper localization. However, the framework embeds watermarks directly in the pixel space, which makes them more susceptible to attacks. Even minor changes in image orientation or scale can significantly alter pixel values. As a result, pixel-based systems become highly sensitive to trivial modifications, even when the semantic content of the image remains unchanged. Consequently, the OmniGuard framework is less effective in detecting deepfakes, where the primary objective is to identify only malicious edits that alter the semantic-level meaning of the image.

In contrast to existing methods, we embed watermarks in the latent space rather than the pixel space, leveraging semantic-level manipulations of the image’s representation. Since the latent space encodes high-level meaning, the watermark becomes tightly coupled with the image’s semantics. Any malicious attempt to alter the image’s meaning disrupts this coupling, effectively breaking the watermark and revealing tampering. Most importantly, through a learnable watermark embedder network, we identify less perceptually salient directions to embed information without significantly altering the human-perceived semantics of the input image. \textcolor{black}{We propose to leverage a spherical latent space from CLIP \cite{radford2021learning} for the embedding of the proposed watermark. CLIP features lie on a unit hypersphere and offer normalised operations for the embedding of our watermark, constraining our watermark perturbations to keep features on this sphere. This prevents the watermark from introducing large deviations in the feature norm, thus preserving image quality. It effectively bounds the perturbation in a way that minor pixel shifts (like rotation, scale) that keep semantics unchanged will not move the feature off the sphere, whereas semantic changes will cause larger movements around the sphere.}

Embedding the watermark in this space offers two key advantages: (1) inherent resilience to benign transformations, such as resizing, compression, or brightness adjustments, that do not significantly alter semantic features, and (2) heightened sensitivity to malicious manipulations that change the semantic meaning of the image, such as identity swaps or expression alterations in deepfakes. 
\textcolor{black}{Latent space embedding reduces artefacts and ties the watermark to semantic content.}

\textcolor{black}{We propose a novel Adversarial Reinforcement Learning (ARL) paradigm for learning watermarks that are both resilient to benign manipulations and fragile to semantic alterations, enabling robust deepfake detection. The framework is structured as an adversarial game between two networks: the watermarker (i.e., the watermark embedder, extractor), and the attacker agent. The watermarking network aims to embed latent space watermarks that survive traditional image transformations but fail under semantic changes. The attacker, in turn, seeks to break the watermark by maximizing extractor loss. It can dynamically adjust the strength and type of attacks, ranging from conventional (e.g., compression, resizing) to semantic (e.g., facial attribute editing via StyleGAN), and combine them into complex, evolving curricula. This adversarial setup forces the watermarker to learn a sophisticated strategy that balances robustness and fragility, avoiding both overly weak and excessively strong watermarking schemes,  allowing us to arrive at the right combination for this paradoxical objective. }The result is an
architecture that achieves unprecedented levels of resilience across bonafide transformations and greater levels of fragility towards malicious manipulations. \textcolor{black}{Fig. \ref{fig:hero_figure} illustrates examples of our method’s performance: row 1 shows genuine images, row 2 shows those images after numerous learned attacks against watermarking were applied. In both cases, the watermark remains intact. Row 3 shows images after malicious semantic edits where the watermark fails, and the system declares them fake.}

The main technical contributions of this paper, through which we introduce the proposed \textit{DeepForgeSeal} framework, can be summarised as follows:
\begin{enumerate}
    \item We introduce a novel deep watermarking framework for deepfake detection that operates in the high-dimensional semantic space of input images.
    \item We design a learnable watermarker with explicit control over message encoding and extraction, achieving semantic stealth.
    \item We propose a new ARL (Adversarial Reinforcement Learning) paradigm that enables the agent to learn a strategy balancing watermark resilience and fragility.
    \item We develop a reward function that guides the watermark attacker to discover failure regions in the latent space and image manipulations that induce significant latent shifts, allowing it to devise a structured curriculum of attacks.
\end{enumerate}

\section{Related Work}

\subsection{Watermarking Approaches for Deepfake Detection}
Passive deepfake detection refers to techniques that analyse the media without requiring any prior knowledge of the generation process or embedding of security features such as watermarks. These approaches have demonstrated remarkable performance over the past decade, primarily by identifying artefacts introduced during the deepfake generation process. Examples include unsynchronised lip movements \cite{haliassos2021lips}, irregular facial landmarks \cite{agarwal2019protecting}, or physiological inconsistencies like abnormal blood oxygen concentration \cite{fernandes2019predicting}. \textcolor{black}{Other forensic methods extend this idea by using deep neural networks to recognise the sequence of image manipulation operations applied to media \cite{liao2020robust}.} However, their effectiveness is inherently tied to the presence of such artefacts, which vary depending on the generation technique used. Consequently, the generalisability of these detectors to novel or unseen deepfake generation methods remains limited. Moreover, the rapid evolution of deepfake technologies poses a significant challenge to maintaining detection robustness, raising concerns about the long-term reliability of passive detection strategies. Such limitations have led to the development of proactive deepfake detection approaches, with watermarking being a prominent example. This technique embeds invisible signals into benign images before any manipulation occurs. These signals act as verification markers, enabling systems to actively determine whether the content has been tampered with based on the presence and integrity of the embedded watermark.

One of the earliest works in watermarking based deepfake detection can be attributed to \cite{wang2021faketagger}, which embeds a robust, invisible watermark into facial images before any manipulation occurs. These watermarks are resilient to various deepfake transformations and image processing operations. The embedded watermark acts as a unique identifier, allowing the origin of a manipulated image to be traced. The works \cite{wang2023robust, wang2024lampmark} have also leveraged the concept of robust watermarking as a line of defense against deepfakes. Specifically, in contrast to \cite{wang2021faketagger}, which embeds identity-linked tags using a learned network which is resilient to deepfake transformations, \cite{wang2023robust, wang2024lampmark} introduce training-free watermarking mechanisms based on facial landmarks. For example, instead of relying on learned embeddings, the authors of \cite{wang2024lampmark} propose to transform structural facial representations into binary perceptual watermarks, enabling robust and imperceptible embedding without extensive model training. \textcolor{black}{Beyond face-specific designs, the broader steganography literature has explored optimising watermark embedding for improved imperceptibility and resilience. Liao et al. \cite{liao2019new, liao2020adaptive} propose adaptive payload partitioning strategies across colour channels and multiple images based on texture complexity, enhancing robustness against statistical detection. In addition, recent work has investigated watermark robustness under extreme real-world distortions; for example, partial screen-shooting watermarking schemes embed redundant watermarks across spatially distributed regions to remain detectable even when only a portion of the image is captured \cite{chen2025flexible}.}

Despite these advances, robust watermarking-based deepfake detection approaches require validating the embedded watermark against a known source or identity, which can be logistically complex in real-world applications, such as in social media platforms. Each image or video must be checked to ensure the watermark matches a trusted source identity, which assumes access to and maintenance of such a database. 

Semi-fragile watermarks, which leverage both the fragility and robustness of watermarking, simplify the detection process by serving as self-contained integrity checks. Specifically, if the watermark is broken or unreadable, it directly signals tampering, without needing to compare against a source. This makes semi-fragile watermarking approaches more efficient and scalable for deployment in environments where rapid and autonomous verification is critical. One of the early works in semi-fragile watermarks for deepfake detection can be attributed to FaceGuard \cite{yang2021faceguard}, which encodes a semi-fragile identity-specific watermark. Along similar lines, Zhao et al. \cite{zhao2023proactive} proposed an identity-based semi-fragile watermarking approach that is sensitive to face swap manipulations while remaining resilient to benign image transformations. Despite its advances, the reliance on identity-specific watermarks for each individual makes these techniques less appealing for real-world applications in which the true identity of the individuals is not available or cannot be readily extracted. WaterLo \cite{beuve2023waterlo} addresses these limitations by introducing a localised, fixed semi-fragile watermark that disappears in manipulated regions, allowing not only detection but also precise localisation of tampered areas. Following this approach, the authors of FaceSigns \cite{neekhara2024facesigns} have introduced a semi-fragile watermarking framework that embeds a 128-bit secret message directly into the image pixels.

Most recently, the EditGuard \cite{zhang2024editguard} introduces a novel approach for decoupling the training process from specific tampering types. Traditional watermarking systems often require retraining or fine-tuning for each manipulation scenario. In contrast, EditGuard’s image-to-image steganography framework generalises across diverse tampering methods, including face swaps, background replacements, and AI-generated edits, without needing retraining. OmniGuard \cite{zhang2025omniguard} further enhances this approach by introducing a hybrid forensic framework that combines proactive embedding of the semi-fragile watermark with passive blind extraction. In a different line of work, the authors of \cite{wang2025fractalforensics} leverage the mathematical properties of fractals to generate watermark patterns through a parameter-driven pipeline. Moreover, watermarks are embedded using an entry-to-patch strategy, where each watermark matrix entry is mapped to a specific image patch, enabling precise localisation of manipulated regions. Though significant strides have been made, the majority of existing semi-fragile watermarking approaches still rely on the pixel space for embedding, rather than leveraging the latent space, which could offer greater flexibility and robustness. Furthermore, current adversarial training pipelines used when training the watermarking frameworks are often simplistic and lack the sophistication needed to simulate complex attack curricula. As a result, the resilience of watermarking techniques under realistic and increasingly sophisticated adversarial conditions remains limited. In contrast, our framework leverages the latent space to obtain more structured and disentangled representations. This enhances watermark embedding consistency, robustness, and sensitivity to manipulations. Additionally, the integration of adversarial reinforcement learning enables the system to simulate and defend against complex, curriculum-based adversarial attacks.
This significantly improves the robustness and generalisability of the watermarking strategy across diverse manipulation scenarios. 

\subsection{Adversarial Reinforcement Learning}

In a learning setting, the relationships among agents can be categorised into cooperative, competitive, and both \cite{wachi2019failure}. However, most of the prior works \cite{foerster2018counterfactual, wang2021multi, peng2021learning, kouzehgar2020multi, wang2021towards} in Multi Agent Reinforcement Learning (MARL) have focused on cooperative tasks, in which the cumulative reward is maximized as a group. Among the limited number of works extending traditional MARL into adversarial settings, researchers have introduced adversarial elements, either as competing agents or perturbations, to enhance the robustness and generalisation capabilities of the overall algorithm. For example, in \cite{li2019robust} the authors have shown the potential of learning a robust generalised policy using Multi Agent Adversarial Reinforcement Learning (MAARL), whereas the authors of \cite{van2020robust} have used MAARL to avoid overfitting. In a similar line of work, Bukharin et. al \cite{bukharin2023robust} address the lack of robustness and sensitivity to environment changes in MARL through adversarial training. Specifically, the authors of \cite{bukharin2023robust} introduce adversarial regularization to enforce Lipschitz continuity in policies, improving robustness against noisy observations. Most recently, Yuan et. al \cite{yuan2024communication} have proposed an approach based on evolutionary learning to enhance robustness in message-passing for improving the communication efficiency of agents. 

While adversarial training has shown promise in both reinforcement learning and watermarking independently, to the best of our knowledge, none of the prior works have investigated the integration of Adversarial Reinforcement Learning (ARL) into watermarking. Through empirical evaluations, we demonstrate that ARL offers a compelling framework for training watermarking agents that can dynamically adapt to image manipulations. This enables watermarking systems that are both robust and semi-fragile, providing resistance to benign transformations while remaining sensitive to malicious tampering.
\section{Methods}

In this section, we discuss our proposed approach. We first provide an overview of the main components that constitute our DeepForgeSeal framework in Sec. \ref{sec:overview}. Our watermarker, watermark attacker, and watermark extraction and deepfake detection processes are discussed in Secs. \ref{sec:watermarking_agent}, \ref{sec:watermark_attacker}, and \ref{sec:watermark_extraction}, respectively. The implementation details of the framework are provided in the supplementary material.

\subsection{Problem Formulation}\label{sec:problem_formulation}

\textcolor{black}{
Let $x$ denote an input image with CLIP semantic feature
$f(x) \in \mathbb{R}^{\zeta}$, $\|f(x)\|_2 = 1$, and let $K$ be a shared secret key from which a watermark message $M$ is deterministically generated. We seek a watermarker $\theta = \{\mu, \phi, \pi\}$ that embeds $M$ in the latent space to produce a watermarked image $x' = \pi(f(x) + p)$, an attacker $\eta$ that maps $x'$ to an adversarial image $x_a$, and an extractor $\delta$ that recovers a message $M'$ from any query image $x''$. The objective is semi-fragility: $M$ must be recoverable under benign transformations (low BER) yet fail under malicious semantic edits (high BER). At verification, integrity is decided by $f_{\text{DEEPFAKE}}(x'') = \big[\, \text{BER}(M', M) > \lambda \,\big]$
(Eq.~\ref{eq:classifier}), where $\lambda$ is a predefined threshold. The full procedure is summarized in Alg. \ref{alg:deepforgeseal}. Furthermore, a summary of notation is provided in Table VI in the supplementary material.}

\subsection{Framework Overview}\label{sec:overview}

Given an input image \( x \), we define a watermarker \( \theta\) that learns the distribution of the semantic features \( f(x) \) in a latent semantic space \( \mathbb{S} \). The watermarker inserts a watermark \( M \) into the latent space, ensuring that it does not significantly alter the semantic meaning of \( x \), and decodes the new image \( x' \) with the watermark from the latent space back to the image space. The objective of the watermarker is to identify the region in $ \mathbb{S}$ where it could embed the watermark without significantly altering the human-perceived meaning of $x$. The watermarker must also learn to maintain an optimal balance between resilience and fragility. Specifically, the embedded watermark should remain robust to benign operations such as JPEG compression, cropping, and color adjustments, while being fragile to malicious edits, such as face swapping or facial attribute editing, that alter the semantic content of the image.

The objective of the watermark attacker agent, \( \eta \), is to generate an adversarial image \( x_a \) from the watermarked image \( x' \), i.e., \( \eta(x') \rightarrow x_a \), by applying transformations that degrade or remove the embedded watermark. To achieve this, the attacker can choose from a range of image manipulations, including benign operations and malicious edits. These manipulations may also be combined, and for each type, the agent can control the associated parameters to vary the strength of the attack. \textcolor{black}{The watermarker is trained via supervised losses, not via an RL algorithm, whereas the attacker agent is trained via reinforcement learning to maximise its reward.}

\begin{figure*}
    \centering
    \includegraphics[width=\linewidth]{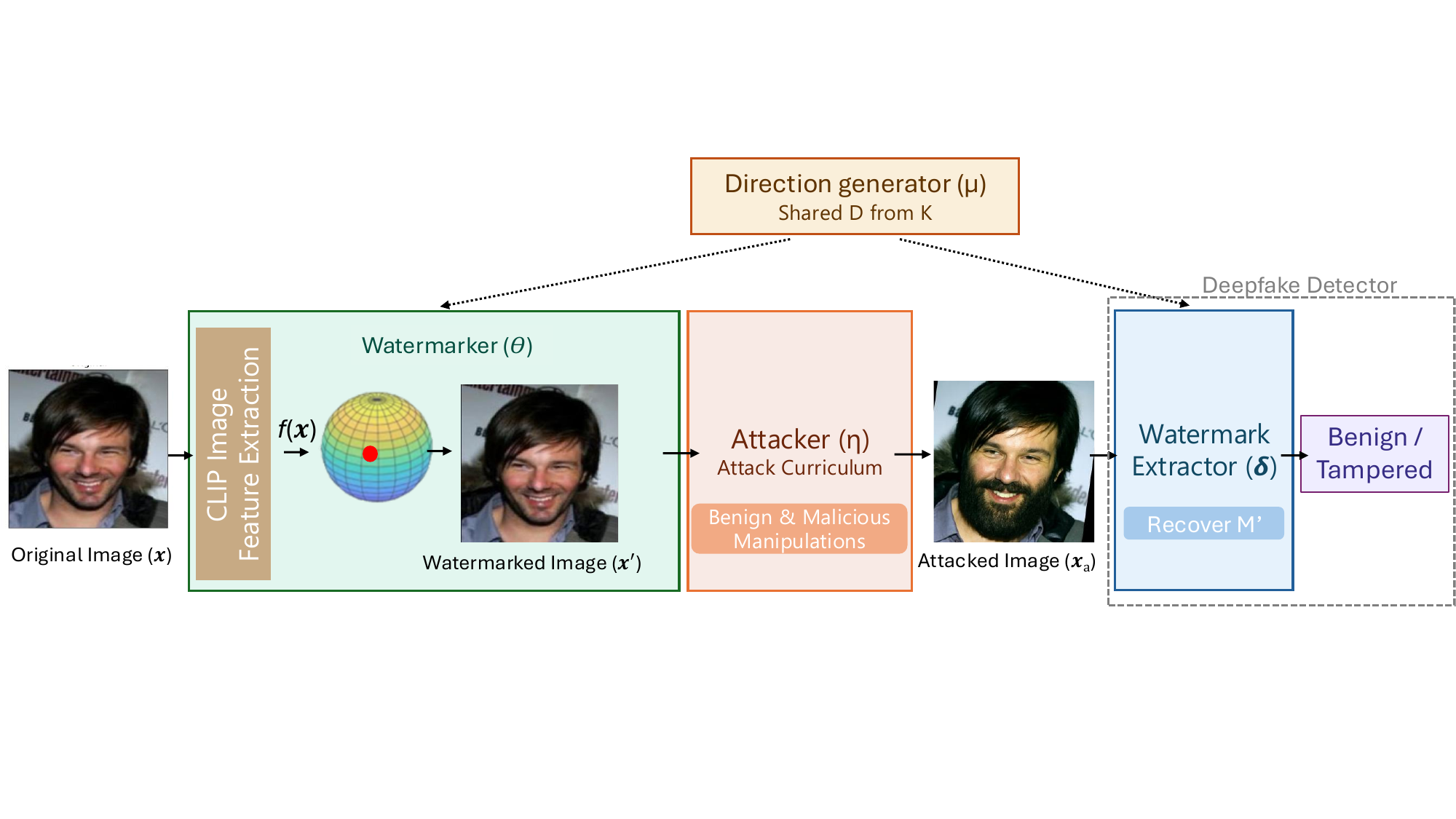}
    \caption{\textcolor{black}{Method Overview: Given an input image \( x \), the watermarker \( \theta \) embeds a watermark \( M \) into the latent space \( \mathbb{S} \) derived from semantic (CLIP Image) features \( f(x) \), producing a watermarked image \( x' \). The direction generator $\mu$ generates this watermark from a secret key K, which is shared by both the watermarker $\theta$ and the extractor $\delta$. An attacker \( \eta \) generates an adversarial image \( x_a\) to disrupt watermark integrity. $\eta$ uses both benign edits (e.g., JPEG compression, cropping) and semantic-altering manipulations (e.g., face swaps) when generating its attack curriculum. The extractor \( \delta \) attempts to recover \( M \) from any image \( x'' \); failure to extract a valid watermark flags \( x'' \) as a potential tampered image, leveraging watermark consistency as a proxy for semantic authenticity.}}
    \label{fig:overview}
\end{figure*}

The watermark extractor, \( \delta \), is responsible for retrieving the embedded watermark from a given image \( x'' \), which may or may not carry a watermark. Since the watermarker \( \theta \) dynamically adjusts the location of the watermark, the extractor must co-adapt and learn in tandem with the watermarker to ensure reliable extraction. In our framework, the success or failure of watermark extraction serves as a heuristic for deepfake detection. Specifically, if the extractor \( \delta \) fails to recover a valid watermark from the image \( x'' \), the image is flagged as a potential deepfake. Our approach leverages the learned consistency between the watermarking and extraction processes as an indirect signal of authenticity, enabling the detection of semantic-level manipulations without relying on a fixed watermark pattern. These modules, and the flow of information between them, are illustrated in Fig. \ref{fig:overview}.

\subsection{Watermarker}\label{sec:watermarking_agent}
The watermarker is responsible for encoding and embedding the watermark into the image. Its architecture is designed to operate within the semantic latent space of a pre-trained CLIP model, ensuring that the embedding process is guided by high-level image representations. The procedure comprises two key stages: (i) directional embedding in the latent sphere, and (ii) perturbative watermark embedding. These stages are discussed in detail in the following sub-sections.

\subsubsection{Directional Embedding in Latent Sphere}
To extract semantically meaningful information from the input image, $x$, our watermarker, $\theta$, operates in the feature space of the CLIP image encoder \cite{radford2021learning}, which is denoted as $E_{\mathrm{CLIP}}$. Specifically, the feature vector $f(x) = E_{\mathrm{CLIP}}(x) \in \mathbb{R}^{\zeta}$ is L2-normalized, such that $\|f(x)\|_2 = 1$, constraining it to the surface of a $\zeta$-dimensional hypersphere. The watermarker has explicit control over the embedding directions that it leverages to embed the message (i.e., watermark), $M$. However, allowing the watermarker to freely modify both the embedded message and the latent space directions makes it infeasible for the watermark extractor to recover the message without access to either the original image \( x \) or the watermark \( M \). To address the challenge of extracting dynamically embedded messages, we propose deriving the latent space directions through a learnable \textbf{direction generator}, $\mu$, while relying on a \textbf{shared secret key} \( K \). \textcolor{black}{The watermark message $M$ is not a fixed bit string but is deterministically generated from a secret key. This design is analogous to cryptographic pseudo-random number generation, where a secret key is used as the seed to produce a reproducible message. Consequently, $M$ does not need to be stored with each image; the secret key alone is sufficient for the detector to regenerate the watermark message during verification.} The key is first projected into the CLIP text embedding space \cite{radford2021learning} to obtain a feature vector \( f(K) \). The direction generator network \( \mu \) then uses this feature to produce a canonical set of \( L \) direction vectors \( D = \{d_1, d_2, \ldots, d_L\} \). Formally,

\textcolor{black}{
\begin{equation}
\begin{aligned}
D &= \operatorname{reshape}\!\left(\mu(f(K))\right) \in \mathbb{R}^{L \times \zeta}, 
\qquad \| d_i \|_2 = 1,\;\forall i, \\
P &= D f(x') \in \mathbb{R}^{L}, 
\qquad P_i = \langle f(x'), d_i \rangle .
\end{aligned}
\label{eq:1}
\end{equation}}

Since \( D \) depends only on \( K \), it remains consistent across all images. However, the directions are still \textbf{learnable}, as the weights \( \omega \) of \( \mu \) are optimised during training. For each bit \( m_i \in \{0, 1\} \) of a dynamically generated message \( M \), the objective is to modify the image \( x \) to produce \( x' \) such that the projection of its new feature vector \( f(x') \) onto each direction \( d_i \) matches a target value,

\begin{equation}
\langle f(x'), d_i \rangle \approx P_i^{\text{target}} =
\begin{cases}
    \xi_1 & \text{if } m_i = 1 \\
    \xi_0 & \text{if } m_i = 0
\end{cases}
\label{eq:2}
\end{equation}

where \( \langle \cdot, \cdot \rangle \) denotes the dot product, and \( \xi_1, \xi_0 \) are predefined projection magnitudes that encode binary values.

\subsubsection{Perturbative Watermark Embedding}
Once the \( m_i \in \{0, 1\} \) of the message \( M \) have been identified, the watermarker employs an MLP layer, $\phi$, that receives semantic features, $f(x)$, of $x$ and the watermark, $M$, and synthesizes a perturbation vector $p$ over $\zeta$-dimensional feature vector $f(x)$ such that:
\begin{equation}
p = \phi([f(x); M]).
\label{eq:3}
\end{equation}
The perturbed latent code is then given by:
\begin{equation}
q = f(x)+p,
\label{eq:4}
\end{equation}
and a decoder \( \pi \) reconstructs the watermarked image \( x' \) from the perturbed representation:
\begin{equation}
x' = \pi(q).
\label{eq:5}
\end{equation}

The watermarker's objective is to minimize a composite loss function $\mathcal{L}_W$ that balances imperceptibility  ($\mathcal{L}_{\text{clip}}$), embedding accuracy ($\mathcal{L}_{\text{dir}}$), and the conditional fragility objective ($\mathcal{L}_{\text{ext}}$). Formally, this can be written as:
\begin{equation}
\mathcal{L}_W = \alpha \mathcal{L}_{\text{clip}} + \beta \mathcal{L}_{\text{dir}} + \gamma \mathcal{L}_{\text{ext}},
\label{eq:6}
\end{equation}

where, 
\begin{align}
\mathcal{L}_{\text{clip}} &= \| f(x) - f(x') \|_2^2 \\
\mathcal{L}_{\text{dir}} &= \text{CrossEntropy}(M, \varpi), \\
\mathcal{L}_{\text{ext}} &= -\text{BER}(M, M'),
\label{eq:lossess}
\end{align}

where $M'$ is the recovered message obtained by decoding the projection vector $P = \{ \langle f(x'), d_i \rangle \}_{i=1}^L$, $\varpi$ denotes predicted message logits by the watermark extractor, from which the final binary message $M'$ is recovered, and $\text{BER}(.,.)$ is Bit Error Rate. See Sec. \ref{sec:watermark_extraction} for details.

\subsection{Watermark Attacker} \label{sec:watermark_attacker}
The objective of the watermark attacker, $\eta$, is to learn an optimal policy to destroy the embedded watermark. One of our key innovations is its ability to compose complex attacks using a predefined set of benign/traditional image transformations and malicious/generative transformations. This enables the agent to learn a dynamic attack curriculum that jointly models both the type and intensity of each attack. The following subsections illustrate: (i) the process of composing a combinatorial attack, (ii) how the watermark attacker agent is trained, and (iii) how it is incentivized to identify attack curricula that lead to substantial semantic drifts and are closer to historically known failure regions in the latent space, where the watermark extractor fails to recover the watermark.

\subsubsection{Composing a Combinatorial Attack}
Let the set of available attacks, including the benign and malicious transformations, be denoted as $\mathcal{A} = \{a_1, \ldots, a_\Gamma\}$. The watermark attacker agent, $\eta$, receives the CLIP features, $f(x')$,  of the watermarked image, $x'$, as input and outputs two vectors: logits for attack selection and normalized strength parameters $\tau$ as:
\begin{equation}
(\text{logits}, \tau) = \eta(f(x')).
\label{eq:10}
\end{equation}
These logits are passed through a sigmoid function (denoted by $\sigma$ to convert them into probabilities:
\begin{equation}
B = \sigma(\text{logits}) \in [0, 1]^\Gamma.
\label{eq:11}
\end{equation} 
Then, for each probability, $B_l$ , we sample a binary action, $a_l$, from a Bernoulli distribution as:  
\begin{equation}
a_l \sim \text{Bernoulli}(B_l) \quad \forall l \in \{1, ..., \Gamma\}
\label{eq:12}
\end{equation} 

The result is a binary action vector $a \in \{0, 1\}^\Gamma$ in which each element is independently sampled, creating a \textbf{combinatorial attack strategy}.

Simultaneously, the strength vector $\tau \in [0, 1]^\Gamma$ determines the intensity for each attack. Each normalized value $\tau_l$ is mapped to a meaningful parameter, $\text{param}_l$, in the selected attack type, $a_l$, and within a predefined range $[\text{min}_l, \text{max}_l]$ as:
\begin{equation}
\text{param}_l = \text{min}_l + ( \text{max}_l - \text{min}_l ) \cdot \tau_l.
\label{eq:13}
\end{equation}

\subsubsection{Learning an Attack Curriculum}
We train our watermark attacker agent using \textbf{reinforcement learning} to learn an optimal policy $\psi$ that maximizes the likelihood of watermark extraction failure, thereby ensuring the success of the attack. Specifically, the primary reward $R$ for the watermark attacker agent can be formulated as the failure of the Watermark Extractor, measured by the Binary Cross-Entropy (BCE) loss between the extracted message $\tilde{M}$ and the original message $M$ as:
\begin{equation}
R_{\text{failure}} = \text{BCE}(\tilde{M}, M).
\label{eq:simple_attack_objective}
\end{equation}
A strong and adaptive attack policy would drive the watermarker to learn more resilient and sophisticated watermarking strategies. In contrast, the simplistic objective in Eq. \ref{eq:simple_attack_objective}, which focused solely on extractor failure, lacks the distinction required to drive meaningful improvements. As such, an advanced reward signal formulation is required to incentivize the attacker to pursue diverse and exploratory policies that provide richer adversarial signals. 

\subsubsection{Promoting Semantic Drift and Targeted Shifts Toward Known Failure Regions} 
We propose to augment the reward with bonuses to encourage exploration and exploitation. Specifically, we define a \textbf{curiosity reward} which is proportional to the squared Euclidean distance between the CLIP features of the image before and after the attack. This could be written as:
\begin{equation}
R_{\text{curiosity}}  = \Delta\|f(x') - f(x_a)\|^2_2,
\end{equation}

where \( x_a\) is the attacked image and  $\Delta$ is a scaling factor. This reward incentivizes the attacker to discover attacks that cause the most significant semantic disruption, which are often the most challenging for a watermark to survive. 

In addition, we introduce a \textbf{proximity reward} that encourages the attacker to perturb the image toward regions of the latent space known to challenge the watermark extractor. In particular, we introduce a memory buffer, $J$, that stores the latent representations, $f(x_a)$, of images that have historically caused failures in watermark extraction. Then, the potential, $\rho$, of a new attacked image $x_a$ can be defined as its proximity to its closest feature vector stored in the failure memory. Now the attacker's proximity reward can be defined as inversely proportional to the distance of its attacked image from the closest failure memory embedding. Formally, this could be written as:

\begin{equation}
\begin{split}
\rho(x_a) &= \min_{f_j \in J} \| f(x_a) - f_j \|_2, \\
R_{\text{proximity}} &= \frac{\nu}{\rho(x_a) + \epsilon}, 
\end{split}
\label{eq:16}
\end{equation}
where $\nu$ is a scaling factor and $\epsilon$ is a small constant to ensure numerical stability.

Now, the final objective of the watermark attacker agent can be written as:
\begin{equation}
\begin{split}
\mathcal{L}_\eta = -\mathbb{E}_{a \sim \psi} \Big[ R_{\text{failure}} + R_{\text{proximity}} + R_{\text{curiosity}} 
- o\sum_{l=1}^\Gamma a_l \Big] \\
- r\sum_{l=1}^\Gamma H(B_l)
\end{split}
\label{eq:17}
\end{equation}
where $\sum_{l=1}^\Gamma a_l$ is a small regularization term penalizing the number of active actions, $H(B_l)$ denotes the entropy of the Bernoulli action distribution for the 
$l^{\textrm{th}}$ attack, and $o$ and $r$ are hyperparameters. It should be noted that $R_{\text{curiosity}}$ encourages goal-directed exploration while $\sum_{l=1}^\Gamma H(B_l)$ is used to encourage policy-level exploration such that the agent maintains diversity in action selection, serving complementary purposes in the learning process. 

\subsection{Watermark Extraction and Deepfake Detection}\label{sec:watermark_extraction}
\textcolor{black}{The watermark extractor, $\delta$, is responsible for recovering the message $M$ from a given image $x''$, which may or may not carry a watermark and could potentially be corrupted.} Specifically, given the input $x''$, it first extracts CLIP features $f(x'')$, and independently generates the direction vectors, $D$, from the secret key, $K$, using the direction generator, $\mu$. Then it produces the projection of $f(x'')$ onto each of the direction vector $d_i \in D$ as:
\begin{equation}
P''_i = \langle f(x''), d_i \rangle.
\label{eq:18}
\end{equation}
The resulting vector of projections $P'' = \{P''_1, ..., P''_L\}$ is passed through an MLP, which outputs the message logits $\varpi$, from which the recovered binary message $M'$ is obtained.

In the proposed framework, the fragility of the watermark serves as the deepfake detection mechanism. Given the direction generator, $\mu$, and the shared secret key $K$, the integrity of an input image can be verified at inference time by computing the Bit Error Rate (BER) between the extracted message $M'$ and the known original message $M$ regenerated from the shared secret key, $K$. \textcolor{black}{We assume that the watermark embedder and the extractor share a secret key $K$ securely, which is a common assumption in watermarking schemes \cite{cao2025secure}.} Since the same trained direction generator $\mu$ and secret key $K$ are used during both watermark embedding and extraction, the resulting direction vectors should remain consistent. A high BER indicates potential generative manipulation, enabling us to leverage the fragility of the watermark as a deepfake detection signal. Formally, this could be written as:
\begin{equation}
f_\text{DEEPFAKE}(x'') = 
\begin{cases}
\text{True}, & \text{if } \text{BER}(M', M) > \lambda \\
\text{False}, & \text{otherwise}
\end{cases}
\label{eq:classifier}
\end{equation}

where $\lambda$ is a predefined classification threshold.

\begin{algorithm}[htbp]
\caption{DeepForgeSeal: Training and Inference}
\label{alg:deepforgeseal}
\small
\begin{algorithmic}[1]
\Require Image set $\mathcal{X}$; secret key $K$; CLIP image/text encoders;
         failure memory $\mathcal{J} \leftarrow \emptyset$
\Require Modules: direction generator $\mu$, perturbation MLP $\phi$,
         decoder $\pi$, extractor $\delta$, attacker $\eta$
\Statex \textbf{\textit{// Training}}
\For{each training iteration}
  \State Sample $x \in \mathcal{X}$; \ $f(x) = E_{\text{CLIP}}(x)$, $\|f(x)\|_2 = 1$
  \State Project $K$ to CLIP-text space to get $f(K)$;
         \ $D = \text{reshape}(\mu(f(K)))$ \Comment{Eq.~\ref{eq:1}}
  \State Generate message $M$ from $K$; set targets $P^{\text{target}}$ \Comment{Eq.~\ref{eq:2}}
  \State $p = \phi([f(x);M])$; \ $q = f(x) + p$; \ $x' = \pi(q)$ \Comment{Eqs.~\ref{eq:3}--\ref{eq:5}}
  \State $(\text{logits}, \tau) = \eta(f(x'))$; \ $B = \sigma(\text{logits})$;
         \ $a_l \sim \text{Bernoulli}(B_l)$ \Comment{Eqs.~\ref{eq:10}--\ref{eq:12}}
  \State Map strengths $\text{param}_l$; apply selected attacks
         sequentially: $x_a = \eta(x')$ \Comment{Eq.~\ref{eq:13}}
  \State Recover message(s) via $\delta$ from $x'$ and $x_a$ \Comment{Eq.~\ref{eq:18}}
  \State Compute $\mathcal{L}_W = \alpha\mathcal{L}_{\text{clip}}
         + \beta\mathcal{L}_{\text{dir}} + \gamma\mathcal{L}_{\text{ext}}$
         \Comment{Eqs.~\ref{eq:6}--\ref{eq:lossess}}
  \State Update $\{\mu, \phi, \pi, \delta\}$ by descending
         $\nabla\mathcal{L}_W$ \Comment{supervised}
  \State Compute $R_{\text{failure}}, R_{\text{curiosity}}, R_{\text{proximity}}$
         \Comment{Eqs.~\ref{eq:simple_attack_objective}--\ref{eq:16}}
  \State Update $\eta$ via REINFORCE on $\mathcal{L}_{\eta}$ \Comment{Eq.~\ref{eq:17}}
  \If{watermark extraction fails on $x_a$}
       add $f(x_a)$ to $\mathcal{J}$
  \EndIf
\EndFor
\Statex \textbf{\textit{// Inference (deepfake detection)}}
\Require Query image $x''$; secret key $K$; threshold $\lambda$
\State $f(x'') = E_{\text{CLIP}}(x'')$; regenerate $D$ and $M$ from $K$
\State $P''_i = \langle f(x''), d_i \rangle$; \ $\delta \rightarrow \varpi \rightarrow M'$
       \Comment{Eq.~\ref{eq:18}}
\State \Return $\textsc{Deepfake} = \big[\, \text{BER}(M', M) > \lambda \,\big]$
       \Comment{Eq.~\ref{eq:classifier}}
\end{algorithmic}
\end{algorithm}

\section{Experiments}
In this section, we present the results of our experiments designed to evaluate and compare the effectiveness of the proposed DeepForgeSeal framework against existing state-of-the-art methods. We begin by detailing the datasets used in our evaluations (Sec. \ref{sec:datasets}), followed by a description of the evaluation metrics employed to assess model performance (Sec. \ref{sec:evalprotocol}). The main experimental results, where we benchmark our method against current state-of-the-art approaches, are provided in Sec. \ref{Sec:main_results}. Additionally, ablation studies conducted to validate the contributions of the learnable directional embedding strategy, adversarial learning pipeline, and novel reward function are discussed in Sec. \ref{Sec:ablations}. For implementation details, hyperparameter evaluations, qualitative evaluations, time complexity analysis, and additional ablation studies, please refer to the supplementary materials.

\subsection{Datasets}
\label{sec:datasets}

\textcolor{black}{Following prior works \cite{zhao2023proactive}, we use the Flickr-Faces-HQ (FFHQ) dataset \cite{karras2019style} to train our model. The CelebA \cite{liu2015deep} and CelebA-HQ \cite{karras2017progressive} datasets are used for evaluation and to demonstrate generalisability. In particular, our method is trained solely on the FFHQ dataset and exposed only to face swapping and attribute‑level manipulations (i.e., change hair color, expression, and age) as defined in \cite{yang2023styleganex}. Therefore, evaluation on the unseen CelebA and CelebA‑HQ datasets demonstrates a strict cross‑dataset generalization.}

The FFHQ dataset contains 70,000 images at a resolution of 1024 × 1024 pixels. In accordance with the dataset authors' guidelines, the first 60,000 images are designated for training, while the remaining 10,000 are reserved for validation. We follow this protocol and use the training subset of FFHQ to train our model. CelebA comprises 10,000 distinct identities, with each identity represented by 20 images at a 128x128 resolution. CelebA-HQ is a high-quality version of the CelebA dataset, consisting of 30,000 images at a resolution of 1024 × 1024 pixels. Specifically, we followed the official test split provided by the CelebA and CelebA-HQ dataset authors to enable direct comparisons. 

\subsection{Evaluation Metics}
\label{sec:evalprotocol}

We evaluate our method from two key perspectives. First, we assess the visual fidelity of the watermarked images compared to the original images using two widely adopted metrics: Peak Signal-to-Noise Ratio (PSNR) and Structural Similarity Index Measure (SSIM). PSNR provides a quantitative measure of the pixel-level differences, where higher values indicate less distortion and better preservation of image quality. SSIM, on the other hand, evaluates perceptual similarity by considering factors such as luminance, contrast, and structural information, offering a more human-aligned assessment of image quality. Together, these metrics help us determine how effectively our method maintains the integrity of the original image while embedding the watermark.

To evaluate the performance of deepfake detection, following prior works \cite{zhao2023proactive}, we compute image-level Accuracy (ACC) and F1-Score. Accuracy reflects the overall proportion of correctly classified images, encompassing both correctly identified real and fake samples. The F1-Score, which balances precision and recall, is particularly informative in capturing the trade-off between false alarms (incorrectly classifying real images as fake) and miss detections (failing to identify fake images). Together, these metrics provide a reliable measure of the system’s ability to detect deepfakes while minimizing classification errors. \textcolor{black}{In addition, for the proposed DeepForgeSeal method and the considered proactive baselines, we report the Area Under the Receiver Operating Characteristic Curve (AUC) metric.}

\subsection{Comparisons with Existing State-of-the-art Methods}\label{Sec:main_results}
In this section, we report results for the proposed model and compare with the existing state-of-the-art methods considering the visual quality of the watermarking (See Sec. \ref{sec:visual_quality}), deepfake detection performance (See Sec. \ref{sec:deepfake_detection_performance}), and watermark's resilience and fragility (See Sec. \ref{sec:watermark_res}). 

\subsubsection{Visual Quality of Watermarking Approaches}\label{sec:visual_quality}
As baselines we use state-of-the-art methods, including FaceSigns  \cite{neekhara2024facesigns}, EditGard \cite{zhang2024editguard}, and Zhao et. al \cite{zhao2023proactive}. When choosing our baselines, we ensured coverage of a diverse range of deep learning approaches.
These include adversarially trained models \cite{neekhara2024facesigns}, dual-purpose watermarking methods for tamper detection and copyright protection \cite{zhang2024editguard}, and identity-entangled watermarking methods \cite{zhao2023proactive}, enabling a comprehensive comparison. 

\begin{table}[htbp]
\centering
\caption{ Visual Fidelity of Different Watermarking Techniques. The best results are highlighted in bold.}
\label{tab:visual_fidelity}
\begin{tabular}{|c|cc|}
\hline
\multirow{2}{*}{Model} & \multicolumn{2}{c|}{Quality Metrics} \\ \cline{2-3} 
                       & \multicolumn{1}{c|}{PSNR ($\uparrow$)}   & SSIM ($\uparrow$)   \\ \hline
 
 FaceSigns   \cite{neekhara2024facesigns}                   & \multicolumn{1}{c|} {32.03}       & 0.92       \\ \hline
 Zhao et. al \cite{zhao2023proactive}                      & \multicolumn{1}{c|}{38.32}      &   0.94     \\ \hline
 EditGard \cite{zhang2024editguard}                      & \multicolumn{1}{c|}{45.07}      &   0.96     \\ \hline
 Omniguard \cite{zhang2025omniguard} & \multicolumn{1}{c|}{46.15} & \textbf{0.97} \\ \hline
 VINE \cite{lurobust}& \multicolumn{1}{c|}{46.83} &  \textbf{0.97} \\ \hline
 DeepForgeSeal (ours) & \multicolumn{1}{c|}{\textbf{48.39}}       &   \textbf{0.97}     \\ \hline
\end{tabular}
\end{table}

Quantitative comparisons for the CelebA-HQ dataset are presented in Tab. \ref{tab:visual_fidelity}. When comparing the proposed DeepForgeSeal with the previous state-of-the-art methods, we observe that our method has the highest PSNR and SSIM, demonstrating our framework’s ability to embed watermarks stealthily, preserving the image’s visual quality. We attribute this to operating within the high-dimensional semantic space of the input images, which enables subtle yet effective modifications. \textcolor{black}{Notably, these results are obtained using only the CLIP feature-similarity loss $L_clip$ to supervise the decoder $\pi$, without any explicit pixel-domain or perceptual reconstruction loss}. In addition to the above quantitative comparisons, we provide qualitative comparisons in the supplementary materials.  

\subsubsection{Deepfake Detection Performance}\label{sec:deepfake_detection_performance}

We compare the proposed DeepForgeSeal method with both passive and proactive deepfake detection methods. Following \cite{zhao2023proactive}, we employ the CelebA and CelebA-HQ datasets to represent low and high-resolution real image cases. Deepfakes are generated using a variety of deepfake generation methods, including AttGAN, StarGAN, InfoSwap
and SimSwap, CIAGAN, and DeepPrivacy. These methods allow us to evaluate the performance of the proposed deepfake detection framework against a wide range of malicious manipulations, including identity swapping and face anonymization techniques. In our experimental pipeline, for passive deepfake detection methods, we use non-watermarked images from the CelebA and CelebA-HQ datasets as real images. The above deepfake generation techniques are then applied to create their fake counterparts. For proactive deepfake detection methods, including the proposed DeepForgeSeal approach, we first watermark the images from the CelebA and CelebA-HQ datasets, and then generate the corresponding fake images using the same deepfake generation methods.

Following \cite{zhao2023proactive}, as baselines for passive deepfake detection we use the state-of-the-art BTS \cite{he2021beyond}, CD \cite{wang2020cnn}, Bonettini et. al \cite{bonettini2021video}, PF \cite{chai2020makes}, RFM \cite{wang2021representative}, SBI \cite{shiohara2022detecting}, FakeShiled \cite{xu2025fakeshield}, and CBO-DD \cite{fernando2025cross}. In addition, the state-of-the-art FaceSigns  \cite{neekhara2024facesigns}, EditGard \cite{zhang2024editguard}, Zhao et al. \cite{zhao2023proactive}, Omniguard \cite{zhang2025omniguard}, and VINE \cite{lurobust}, which are proactive deepfake detection methods, have also been used for comprehensiveness. \textcolor{black}{Results of all passive baselines except FakeShiled \cite{xu2025fakeshield}, and CBO-DD are obtained from Zhao et al. \cite{zhao2023proactive} under their stated evaluation protocol, i.e., using pre-trained models provided by the authors of the passive baseline models, without retraining or threshold tuning on the test data. FakeShiled and CBO-DD models were re-trained using the exact hyperparameters and thresholds provided by the authors to ensure reproducibility of the results. For all proactive deepfake detection baseline methods, we re-evaluated the baselines using the provided public implementations and the thresholds.}

\begin{table*}[htbp]
\centering
\caption{\textbf{Accuracy ($\uparrow$)} and \textbf{F1 Scores ($\uparrow$)} of different methods’ DeepFake detection results. The best results are highlighted in bold.}
\resizebox{\linewidth}{!}{%
\renewcommand{\arraystretch}{1.2}
\setlength{\tabcolsep}{3.5pt}
\begin{tabular}{l|cccccc|cccccc}
\hline
\multirow{2}{*}{\textbf{Detection methods}} & 
\multicolumn{6}{c|}{\textbf{Low Resolutions (CelebA)}} & 
\multicolumn{6}{c}{\textbf{High Resolutions (CelebA-HQ)}} \\ \cline{2-13}
& AttGAN & CIAGAN & DeepPrivacy & InfoSwap & SimSwap & StarGAN2 & 
AttGAN & CIAGAN & DeepPrivacy & InfoSwap & SimSwap & StarGAN2 \\ \hline
BTS~\cite{he2021beyond} & 0.86/0.87 & 0.51/0.66 & 0.50/0.66 & 0.49/0.66 & 0.51/0.67 & 0.53/0.67 &
0.86/0.87 & 0.50/0.66 & 0.50/0.66 & 0.49/0.65 & 0.50/0.66 & 0.55/0.68 \\
CD~\cite{wang2020cnn} & 0.88/0.86 & 0.51/0.03 & 0.51/0.01 & 0.54/0.17 & 0.51/0.01 & 0.78/0.71 &
0.81/0.77 & 0.51/0.04 & 0.52/0.07 & 0.52/0.07 & 0.52/0.06 & 0.84/0.81 \\
Bonettini et. al \cite{bonettini2021video} & 0.59/0.69 & 0.62/0.62 & 0.58/0.68 & 0.57/0.64 & 0.60/0.71 & 0.63/0.63 &
0.53/0.68 & 0.65/0.62 & 0.56/0.69 & 0.51/0.66 & 0.55/0.69 & 0.49/0.65 \\
PF~\cite{chai2020makes} & 0.76/0.79 & 0.51/0.66 & 0.52/0.65 & 0.57/0.68 & 0.54/0.67 & 0.99/0.98 &
0.75/0.79 & 0.51/0.66 & 0.55/0.68 & 0.56/0.69 & 0.54/0.68 & \textbf{0.98/0.97} \\
RFM~\cite{wang2021representative} & 0.50/0.67 & 0.51/0.67 & 0.51/0.67 & 0.50/0.67 & 0.51/0.67 & 0.50/0.67 &
0.50/0.67 & 0.50/0.67 & 0.51/0.67 & 0.50/0.67 & 0.50/0.67 & 0.50/0.67 \\
SBI~\cite{shiohara2022detecting} & 0.79/0.82 & 0.77/0.8 & 0.78/0.82 & 0.77/0.81 & 0.78/0.81 & 0.72/0.75 &
0.80/0.80 & 0.72/0.78 & 0.78/0.80 & 0.76/0.77 & 0.83/0.84 & 0.69/0.70 \\ 
FakeShield \cite{xu2025fakeshield} & 0.84/0.83 & 0.92/0.93 & 0.85/0.86 & 0.78/0.85  & 0.84/0.87  &  0.83/0.83 & 0.83/0.85
 &  0.91/0.93 & 0.84/0.84 & 0.81/0.83 & 0.88/0.90 & 0.81/0.83 \\ 
CBO-DD \cite{fernando2025cross} & 0.84/0.84 & 0.91/0.93 & 0.86/0.87 & 0.79/0.85  & 0.84/0.88  &  0.85/0.85 & 0.83/0.86
 &  0.92/0.93 & 0.86/0.87 & 0.80/0.85 & 0.89/0.91 & 0.82/0.84 \\ \hline
FaceSigns  \cite{neekhara2024facesigns} & 0.91/0.93 & 0.83/0.84 & 0.91/0.95  & 0.93/0.95  & 0.95/0.95 & 0.96/0.98 & 0.87/0.88
 & 0.92/0.93  & 0.80/0.83  &  0.91/0.91 & 0.82/0.82 &  0.84/0.87\\ 
Zhao et. al \cite{zhao2023proactive} & 0.94/0.94 & 0.87/0.86 & 0.98/0.98 & 0.98/0.98 & 0.97/0.98 & 0.82/0.84 &
0.94/0.94 & 0.85/0.82 & 0.99/0.98 & \textbf{0.99/0.99} & 0.98/0.98 & 0.85/0.87 \\
EditGard \cite{zhang2024editguard} & 0.96/0.97 & 0.88/0.88  & 0.96/0.98  &  0.94/0.97 & 0.97/0.98  & 0.91/0.93 & 0.95/0.96
 & 0.87/0.91  & 0.97/0.99 & \textbf{0.99/0.99} & 0.98/0.99  & 0.93/0.95 \\ 
 Omniguard \cite{zhang2025omniguard} & 0.96/0.97 & 0.87/0.87  & 0.96/0.98  &  0.93/0.91 & 0.97/0.98  & 0.92/0.94 & 0.95/0.96 & 0.88/0.92  & 0.97/0.98 & \textbf{0.99/0.99} & \textbf{0.99/0.99}  & 0.94/0.95 
 \\ 
 VINE \cite{lurobust} & 0.95/0.96 & 0.89/0.88  & 0.97/0.98  &  0.95/0.96 & 0.97/0.97  & 0.96/0.97 & 0.96/0.97 & 0.88/0.92 & 0.98/0.99 & 0.98/0.99 & \textbf{0.99/0.99}  & 0.95/0.95 
 \\ 
\hline
\textbf{DeepForgeSeal (Ours)} & \textbf{0.96/0.98} & \textbf{0.92/0.95} & \textbf{0.99/0.99} & \textbf{0.98/0.99} & \textbf{0.98/0.98} & \textbf{0.99/0.99} &
\textbf{0.96/0.98} & \textbf{0.95/0.96}& \textbf{0.99/0.99} & \textbf{0.99/0.99} & \textbf{0.99/0.99} & \textbf{0.98/0.97} \\ 
\hline
\end{tabular}}
\label{tab:deepfake_detection}
\end{table*}

Tab. \ref{tab:deepfake_detection} summarises the comparison results for deepfake detection in CelebA and CelebA-HQ datasets. It is evident that passive deepfake detection methods generally struggle to generalise across different deepfake generation techniques. Among the approaches evaluated, only methods such as SBI~\cite{shiohara2022detecting} and CBO-DD~\cite{fernando2025cross} consistently demonstrate strong performance across all considered generation methods. In contrast, most other passive detection techniques perform well against a limited subset of generation methods but show significant performance degradation when applied to others. This limitation stems from their training approach, which focuses on identifying artefacts specific to the generation techniques used during training. Consequently, these methods tend to perform poorly when tested on deepfakes produced by unfamiliar or diverse generation techniques.

When comparing the results of the proactive detection methods FaceSigns  \cite{neekhara2024facesigns}, Zhao et. al \cite{zhao2023proactive}, EditGard \cite{zhang2024editguard}, Omniguard \cite{zhang2025omniguard}, and VINE \cite{lurobust} with the proposed DeepForgeSeal method, we see that our method has been able to achieve significant detection accuracy across all considered generation methods. Although state-of-the-art methods such as Zhao et. al \cite{zhao2023proactive},  EditGard \cite{zhang2024editguard}, Omniguard \cite{zhang2025omniguard}, and VINE \cite{lurobust} have been able to achieve impressive performance against deepfake generation methods such as AttGAN and DeepPrivacy, their performance degrades when tested against DeepPrivacy and StarGAN2 generation methods. In contrast, our proposed DeepForgeSeal method consistently demonstrates superior detection performance across all evaluated deepfake generation techniques. We attribute this to the learnable watermarking strategy embedded within the framework, which enables explicit control over message encoding and extraction. Additionally, the adversarial reinforcement learning pipeline allows the watermarking agent to actively observe and respond to malicious attacks, dynamically adapting its watermarking strategy. Together, these components contribute to DeepForgeSeal's strong generalisation capability across a wide range of deepfake generation methods. \textcolor{black}{In addition to these comparisons, we compare the proposed method and the considered proactive baselines using the AUC metric, which is available in the supplementary material.}

\subsubsection{Watermark Resilience and Fragility Against Attacks}\label{sec:watermark_res}

In this section, we compare the resilience of our DeepForgeSeal watermarking methods with the current state-of-the-art methods under different attacks against watermarks. For comprehensiveness, we consider both traditional benign attacks, such as cropping, compression, and color changes, as well as the learnable benign attacks proposed by Zhao et. al \cite{zhao2024invisible} and Lukas et. al \cite{lukas2023leveraging}. In addition, to quantitatively assess the fragility against malicious image manipulations, we generate attacks using state-of-the-art GAN-based face swapping and reenactment techniques, including SimSwap \cite{simswap}, UniFace \cite{xu2022designing}, FaceDancer \cite{rosberg2023facedancer}, and HyperReenact \cite{bounareli2023hyperreenact}.  

\begin{table*}[htbp]
\centering
\caption{Bit recovery accuracy (BRA) of baseline FaceSigns \cite{neekhara2024facesigns} and EditGard \cite{zhang2024editguard} watermarking methods and proposed DeepForgeSeal method under benign and malicious transformations. The best results are highlighted in bold.}
\resizebox{\linewidth}{!}{%
\begin{tabular}{|c|ccccccc|cccc|}
\hline
\multirow{2}{*}{Method} & \multicolumn{7}{c|}{BRA (\%) - Benign Transforms ($\uparrow$)}                                                                                                                                               & \multicolumn{4}{c|}{BRA (\%) - Malicious Transforms ($\downarrow$)}                                                         \\ \cline{2-12} 
                        & \multicolumn{1}{c|}{JPG} & \multicolumn{1}{c|}{Noise} & \multicolumn{1}{c|}{Crop} & \multicolumn{1}{c|}{Jitter} & \multicolumn{1}{c|}{Affine} & \multicolumn{1}{c|}{Zhao et. al \cite{zhao2024invisible}} & Lukas et. al \cite{lukas2023leveraging} & \multicolumn{1}{c|}{SimSwap} & \multicolumn{1}{c|}{UniFace} & \multicolumn{1}{c|}{FaceDancer} & HyperReenact \\ \hline
FaceSigns  \cite{neekhara2024facesigns}             & \multicolumn{1}{c|}{0.87}    & \multicolumn{1}{c|}{0.73}      & \multicolumn{1}{c|}{0.72}     & \multicolumn{1}{c|}{0.91}       & \multicolumn{1}{c|}{0.96}       & \multicolumn{1}{c|}{0.87}            &   0.88           & \multicolumn{1}{c|}{0.52}        & \multicolumn{1}{c|}{0.49}        & \multicolumn{1}{c|}{0.62}           & 0.51             \\ \hline
EditGard  \cite{zhang2024editguard}              & \multicolumn{1}{c|}{0.95}    & \multicolumn{1}{c|}{0.78}      & \multicolumn{1}{c|}{0.96}     & \multicolumn{1}{c|}{0.96}       & \multicolumn{1}{c|}{0.99}       & \multicolumn{1}{c|}{0.92}            &      0.95        & \multicolumn{1}{c|}{0.48}        & \multicolumn{1}{c|}{0.49}        & \multicolumn{1}{c|}{0.44}           & 0.50             \\ \hline
Omniguard \cite{zhang2025omniguard}             & \multicolumn{1}{c|}{0.96}    & \multicolumn{1}{c|}{0.83}      & \multicolumn{1}{c|}{0.96}     & \multicolumn{1}{c|}{0.98}       & \multicolumn{1}{c|}{\textbf{1.0}}       & \multicolumn{1}{c|}{0.95}            &      0.96        & \multicolumn{1}{c|}{0.36}        & \multicolumn{1}{c|}{0.28}        & \multicolumn{1}{c|}{0.25}           & 0.42             \\ \hline
VINE \cite{lurobust}              & \multicolumn{1}{c|}{0.96}    & \multicolumn{1}{c|}{0.81}      & \multicolumn{1}{c|}{0.97}     & \multicolumn{1}{c|}{0.95}       & \multicolumn{1}{c|}{0.99}       & \multicolumn{1}{c|}{0.94}            &      0.98        & \multicolumn{1}{c|}{0.31}        & \multicolumn{1}{c|}{0.25}        & \multicolumn{1}{c|}{0.29}           & 0.48             \\ \hline
DeepForgeSeal (ours)                   & \multicolumn{1}{c|}{\textbf{1.00}}    & \multicolumn{1}{c|}{\textbf{0.98}}      & \multicolumn{1}{c|}{\textbf{0.98}}     & \multicolumn{1}{c|}{\textbf{0.99}}       & \multicolumn{1}{c|}{\textbf{1.00}}       & \multicolumn{1}{c|}{\textbf{0.98}}            &  \textbf{0.99}           & \multicolumn{1}{c|}{\textbf{0.24}}        & \multicolumn{1}{c|}{\textbf{0.11}}        & \multicolumn{1}{c|}{\textbf{0.10}}           &  \textbf{0.06}            \\ \hline
\end{tabular}}
\label{tab:attack_resilence}
\end{table*}

The evaluation results are presented in Tab. \ref{tab:attack_resilence}. As baselines, we use the state-of-the-art  FaceSigns \cite{neekhara2024facesigns} and EditGard \cite{zhang2024editguard} semi-fragile watermarking methods. Following prior works \cite{neekhara2024facesigns, zhang2024editguard}, we measure the Bit Recovery Accuracy (BRA) as the evaluation metric, where we expect a higher BRA against benign and lower BRA against malicious transforms to achieve the goal of semi-fragile watermarking. As reported in Tab. \ref{tab:attack_resilence}, we observe that both baseline models perform poorly against benign image manipulations, demonstrating that these watermarking techniques are too fragile to withstand the benign image transformations that users can apply in the real world. At the same time, both baseline techniques demonstrate less fragility towards the malicious, content-altering manipulations that the considered face swapping and reenactment techniques have generated. These evaluations clearly demonstrate the limitations of the existing state-of-the-art watermarking techniques and the superiority of the  DeepForgeSeal paradigm that enables us to learn a strategy that optimally balances watermark resilience and fragility. 

\subsection{Ablation Evaluations}\label{Sec:ablations}
We conducted a series of ablation studies to systematically analyse the impact of the individual innovations that our DeepForgeSeal framework proposes. Several design choices contribute to the robustness of our model: i) the proposed learnable directional embedding strategy that uses a latent sphere; ii) the proposed adversarial learning pipeline with composed attacks with a learnable attack curriculum; and iii) the novel reward function that promotes semantic drift and targeted shifts toward known failure regions. All ablation experiments were conducted on the CelebA-HQ dataset, and for testing the ablation models, we used the validation set of CelebA-HQ. As evaluation metrics, we use PSNR and average BRA against benign and malicious manipulations.  

\subsubsection{Effects of Learnable Directional Embedding in Latent Sphere}
To study the effect of the proposed learnable directional embedding strategy that uses a latent sphere, we generated two ablation variants of the proposed DeepForgeSeal model: i) DeepForgeSeal - w/o [D] - a model with naive latent embedding in which we embed the message directly in the latent space without directional encoding, and ii) DeepForgeSeal - FD - a model with fixed directional embeddings. 

\begin{table}[htbp]
\caption{Effect of Learnable Directional Embedding in Latent Sphere. The best results are highlighted in bold.}
\label{tab:direction_embedding_ablation}
\resizebox{\linewidth}{!}{%
    \begin{tabular}{|c|c|c|c|c|}
        \hline
        Model & PSNR ($\uparrow$) & SSIM ($\uparrow$)  & BRA - Benign ($\uparrow$) & BRA - Malicious ($\downarrow$) \\
        \hline
        DeepForgeSeal - w/o [D] & 46.23 & 0.95 & 0.78 & 0.42 \\\hline
        DeepForgeSeal - FD & 46.58 & 0.95 & 0.86 & 0.19 \\\hline
        DeepForgeSeal & \textbf{48.12} & \textbf{0.97}  & \textbf{0.99} & \textbf{0.12}\\
        \hline
    \end{tabular}}
\end{table}

Results of this comparison are shown in Tab. \ref{tab:direction_embedding_ablation}. We observe that the learnable directional embedding strategy is an integral part of the proposed framework. Specifically, we observe that the naive latent embedding approach, where the message is directly embedded into the latent space, leads to a reduction in PSNR, distorts the natural image characteristics, and diminishes the watermark's resilience against benign image transformations. The introduction of directional embedding enhances the robustness of the watermark (see the row corresponding to DeepForgeSeal - FD). Notably, the experimental results demonstrate that the learnable directional embedding strategy, which encodes the message within a latent sphere, achieves the most favorable balance between resilience and fragility, while preserving image fidelity. These experiments confirm the utility of our learnable directional embedding procedure.

\subsubsection{Effects of Composing a Combinatorial Attack and Learning an Attack Curriculum}





The effectiveness of the proposed adversarial learning pipeline with composed attacks with a learnable attack curriculum is evaluated in this experiment. We generated eight ablation variants of the proposed DeepForgeSeal model: i) DeepForgeSeal - w/o $\eta$ - a model without a watermark attacker agent, ii) DeepForgeSeal - w/o $\mathcal{A}$ - a model in which one fixed attack type (no composition or learning of an attack curriculum) is applied, iii) DeepForgeSeal - $\mathcal{|A|}$ -  a model in which multiple attacks are applied in a fixed sequence, iv) DeepForgeSeal - $\mathcal{\tilde{A}}$ -  a model in which multiple random combinations of attacks applied, v) DeepForgeSeal - $a_1, \tau^\uparrow$ - a model in which a single attack type with gradually increasing strength is applied, vi) DeepForgeSeal - $\mathcal{A^\rightarrow}$ - a model which progressively introduces more complex combinations of attacks, vii) DeepForgeSeal - $a_1, \tau$ - a model with single attack type with learnable strength parameter, viii) DeepForgeSeal - [$\mathcal{A}$] - a model with a predefined attack curriculum without adaptation to agent performance.

\begin{table}[htbp]
\caption{Effects of Composing a Combinatorial Attack and Learning an Attack Curriculum. The best results are highlighted in bold.}
\label{tab:ablation_2}
\resizebox{\linewidth}{!}{%
    \begin{tabular}{|c|c|c|c|c|}
        \hline
        Model & PSNR ($\uparrow$) & SSIM ($\uparrow$) & BRA - Benign ($\uparrow$) & BRA - Malicious ($\downarrow$) \\
        \hline
        DeepForgeSeal - w/o $\eta$ & 32.11 & 0.92  & 0.61 &  0.59\\\hline
        DeepForgeSeal - w/o $\mathcal{A}$ & 41.58 &  0.93 & 0.72 & 0.55  \\\hline
        DeepForgeSeal - $\mathcal{|A|}$ & 46.20 & 0.95 & 0.82 & 0.37  \\\hline
        DeepForgeSeal - $\mathcal{\tilde{A}}$ & 47.15 & 0.96 & 0.87  & 0.31 \\\hline
        DeepForgeSeal - $a_1, \tau^\uparrow$ & 47.12 & 0.96 & 0.76 &  0.42 \\\hline
        DeepForgeSeal - $\mathcal{A^\rightarrow}$ & 48.09 & 0.96 & 0.92 & 0.22 \\\hline
        DeepForgeSeal - $a_1, \tau$ & 48.05 & 0.96  & 0.79 &  0.38\\\hline
        DeepForgeSeal - [$\mathcal{A}$] & 47.11 & 0.95&  0.89 & 0.28  \\\hline
        
        DeepForgeSeal & \textbf{48.12} & \textbf{0.97} & \textbf{0.99} & \textbf{0.12} \\
        \hline
    \end{tabular}}
\end{table}

From the results in Tab. \ref{tab:ablation_2}, we can confirm that there are benefits to incorporating both combinatorial attacks with a learnable attack curriculum as an adversary in the proposed framework, which is evidenced by the rows in Tab. \ref{tab:ablation_2} corresponding to models DeepForgeSeal - $\mathcal{A^\rightarrow}$, and DeepForgeSeal - $a_1, \tau$. Most importantly, when the attacker agent has the knowledge of the current vulnerabilities of the victim model (i.e. watermarking agent), it helps the attacker dynamically adapt its strategy and generate a more complex curriculum of attacks, in turn helping the victim to understand the vulnerabilities of its watermarking strategy and rectifying those. We believe this is the reason for the poor performance of the DeepForgeSeal - [$\mathcal{A}$] model, in which a predefined attack curriculum was used without adaptation to the victim's performance. Furthermore, rows corresponding to DeepForgeSeal - $\mathcal{|A|}$, and DeepForgeSeal - $\mathcal{\tilde{A}}$ confirm the importance of using a combinatorial attack in which a learnable agent determines the attack sequence. Therefore, using this experiment, we can confirm the necessity of both combinatorial attacks and learning an attack curriculum to enhance the robustness of the watermark, while simultaneously preserving image fidelity.

\subsubsection{Effects of Promoting Semantic Drift and Targeted Shifts Toward Known Failure Regions}

To better establish the contributions of the proposed reward function that promotes semantic drift and targeted shifts toward known failure regions, we conducted a further ablation experiment. Specifically, three ablation variants of the proposed model were generated: i) DeepForgeSeal - w/o [$R_{\text{proximity}}, R_{\text{curiosity}}$] - proposed model without both curiosity reward and proximity reward, ii) DeepForgeSeal - w/o $R_{\text{proximity}}$ - proposed model with curiosity reward but without proximity reward, iii) DeepForgeSeal - w/o $R_{\text{curiosity}}$ - proposed model without curiosity reward but with proximity reward. 

\begin{table}[htbp]
\caption{Effects of the Promoting Semantic Drift and Targeted Shifts Toward Known Failure Regions. The best results are highlighted in bold.}
\label{tab:ablation_rewards}
\resizebox{\linewidth}{!}{%
    \begin{tabular}{|c|c|c|c|c|}
        \hline
        Model & PSNR ($\uparrow$) & SSIM ($\uparrow$) & BRA - Benign ($\uparrow$) & BRA - Malicious ($\downarrow$) \\
        \hline
        DeepForgeSeal - w/o [$R_{\text{proximity}}, R_{\text{curiosity}}$] & 48.10 & 0.96  & 0.82 & 0.35 \\\hline
        DeepForgeSeal - w/o $R_{\text{proximity}}$ & 48.12 & 0.96 &  0.91 & 0.24 \\\hline
        DeepForgeSeal - w/o $R_{\text{curiosity}}$ & 48.08 & 0.96 & 0.94  & 0.21  \\\hline
        DeepForgeSeal & \textbf{48.12} & \textbf{0.97} & \textbf{0.99} & \textbf{0.12}\\
        \hline
    \end{tabular}}
\end{table}

The results presented in Tab. \ref{tab:ablation_rewards} confirm the need for the proposed innovative reward function. In particular, the ablation model without both the proximity and the curiosity rewards (see the row corresponding to DeepForgeSeal - w/o [$R_{\text{proximity}}, R_{\text{curiosity}}$] in Tab. \ref{tab:ablation_rewards}) struggles to achieve a good watermark robustness. This is because the attacker agent has learned a sub-optimal attacking policy, marking the watermarking agent unprepared for real-world attacks against watermarks. Comparing rows corresponding to DeepForgeSeal - w/o $R_{\text{proximity}}$ and DeepForgeSeal - w/o $R_{\text{curiosity}}$ models in Tab. \ref{tab:ablation_rewards}, we can confirm that proximity and curiosity rewards complement each other. By leveraging underexplored regions of the latent space alongside known failure zones encountered during watermark retrieval, the attacker agent can formulate a sophisticated attack policy. This, in turn, drives the watermarker agent to refine and strengthen its watermarking strategy. These observations strongly validate the importance of the proposed  reward function that promotes semantic drift and targeted shifts toward known failure regions


For details of hyperparameter evaluation, sensitivity analysis, qualitative evaluations, time-complexity and parameter-count comparison against FaceSigns and EditGuard, and a discussion on limitations of \textit{DeepForgeSeal}, please refer to the supplementary material. 
\section{Conclusion}
This paper presented \textit{DeepForgeSeal}, a Latent Space-Driven Semi-Fragile Watermarking using Adversarial Reinforcement Learning (ARL) for accurate detection of face deepfakes. We demonstrate that the proposed learnable watermarking agent achieves enhanced robustness and semantic stealth through a directional embedding strategy in the latent space. The ARL paradigm enabled the agent to dynamically balance resilience and fragility by interacting with a curriculum of benign and adversarial manipulations introduced by an attacker agent. To intensify this adversarial interaction, we designed reward functions that encouraged semantic drift and targeted perturbations toward known failure regions. This incentivised the attacker agent to develop increasingly sophisticated attack strategies.
In turn, this drove the watermarking agent to refine and strengthen its embedding approach, resulting in a more resilient and adaptive watermarking system. Extensive experiments were conducted on two public benchmarks: CelebA and CelebA-HQ, which demonstrated the ability of the proposed framework to outperform the current state-of-the-art algorithms by significant margins.

\section*{Acknowledgment}
The research was supported by the Australian Government through the Office of National Intelligence Postdoctoral Grant awarded to the primary author under Project NIPG-2024-022.

\bibliographystyle{IEEEtran}
\bibliography{IEEEabrv,Bibliography}
\vspace{-1 cm}
\begin{IEEEbiography}[{\includegraphics[width=1in,height=1.25in,clip,keepaspectratio]{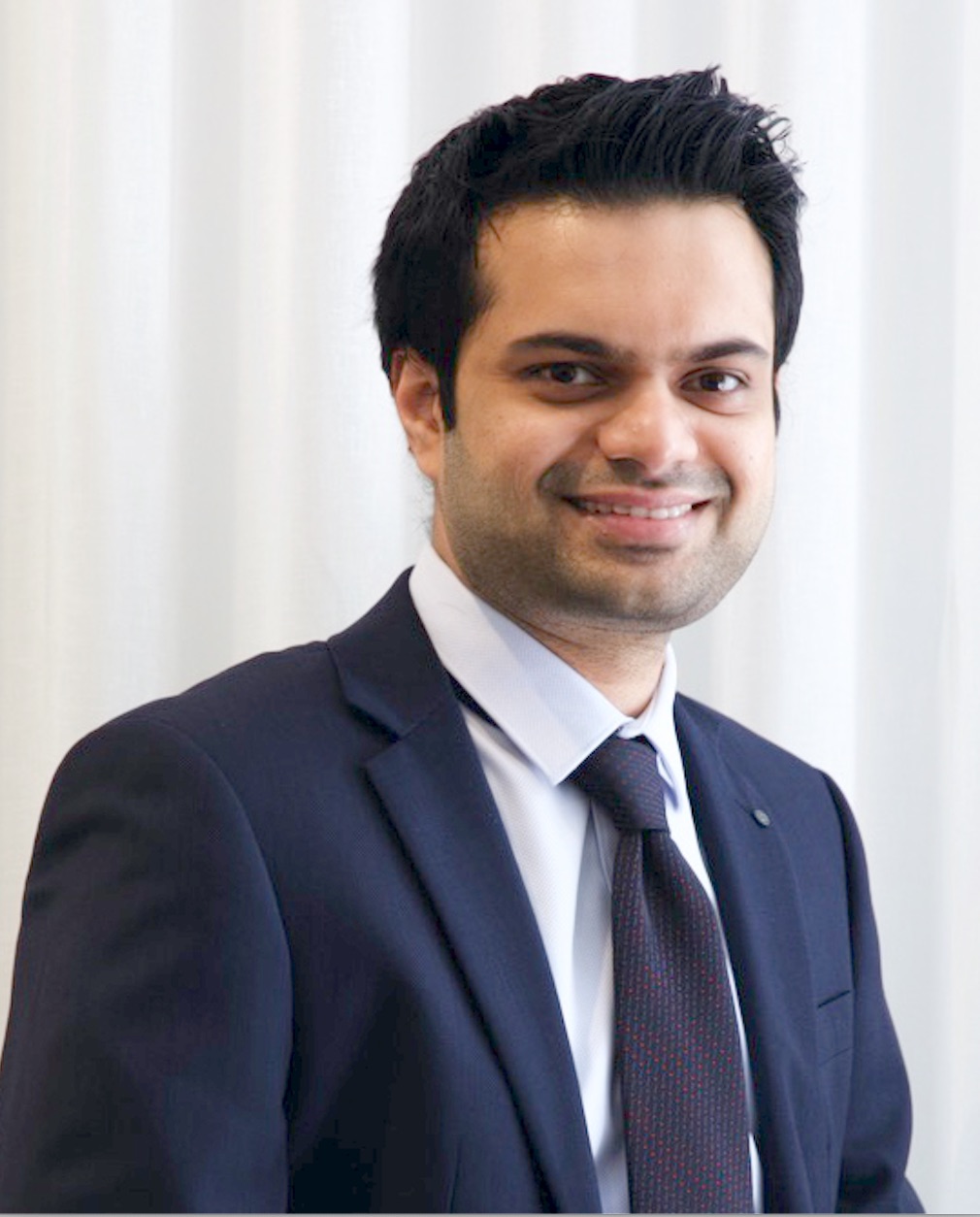}}]{Tharindu Fernando } received his BSc (special degree in computer science) from the University of Peradeniya, Sri Lanka, and his PhD from Queensland University of Technology (QUT), Australia. He is currently a Postdoctoral Research Fellow in the Signal Processing, Artificial Intelligence, and Vision Technologies (SAIVT) research program at the School of Electrical Engineering and Robotics at Queensland University of Technology (QUT). He is a recipient of the 2019 QUT University Award for Outstanding Doctoral Thesis and the 2024 National Intelligence Post-Doctoral Grant.
\end{IEEEbiography}
\vspace{-.5cm}
\begin{IEEEbiography}[{\includegraphics[width=1in,height=1.25in,clip,keepaspectratio]{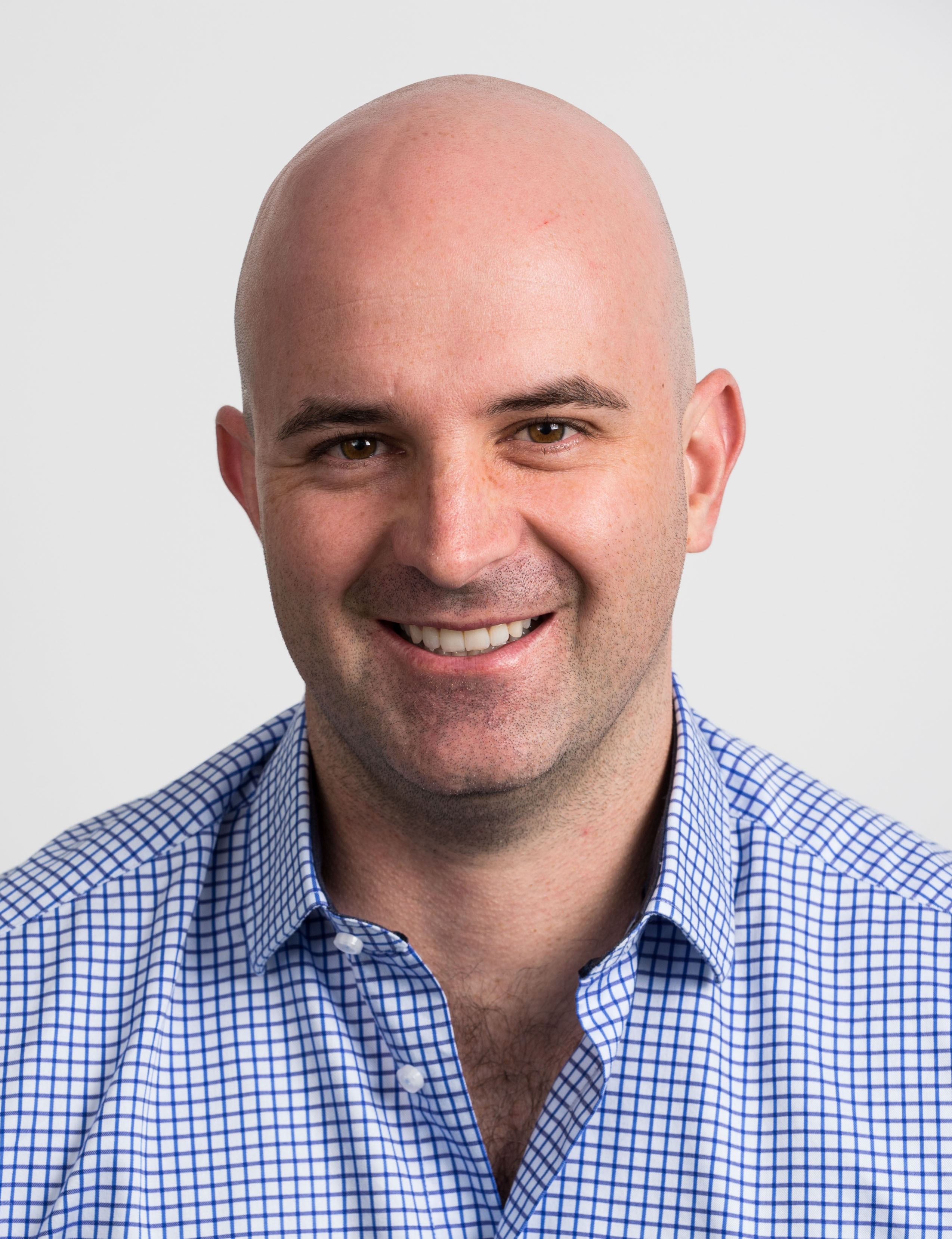}}]{Clinton Fookes}{\space}(Senior Member, IEEE) received the B.Eng. in Aerospace/Avionics, the MBA degree, and the Ph.D. degree in computer vision. He is currently the Associate Dean Research, a Professor of Vision and Signal Processing, and Co-Director of the SAIVT Lab (Signal Processing, Artificial Intelligence and Vision Technologies) with the Faculty of Engineering at the Queensland University of Technology, Brisbane, Australia. His research interests include computer vision, machine learning, signal processing, and artificial intelligence. He serves on the editorial boards for IEEE TRANSACTIONS ON IMAGE PROCESSING and Pattern Recognition. He has previously served on the Editorial Board for IEEE TRANSACTIONS ON INFORMATION FORENSICS AND SECURITY. He is a Fellow of the International Association of Pattern Recognition, a Fellow of the Australian Academy of Technological Sciences and Engineering, and a Fellow of the Asia-Pacific Artificial Intelligence Association. He is a Senior Member of the IEEE and a multi-award winning researcher including an Australian Institute of Policy and Science Young Tall Poppy, an Australian Museum Eureka Prize winner, Engineers Australia Engineering Excellence Award, Australian Defence Scientist of the Year, and a Senior Fulbright Scholar.
\end{IEEEbiography}

\begin{IEEEbiography}[{\includegraphics[width=1in,height=1.25in,clip,keepaspectratio]{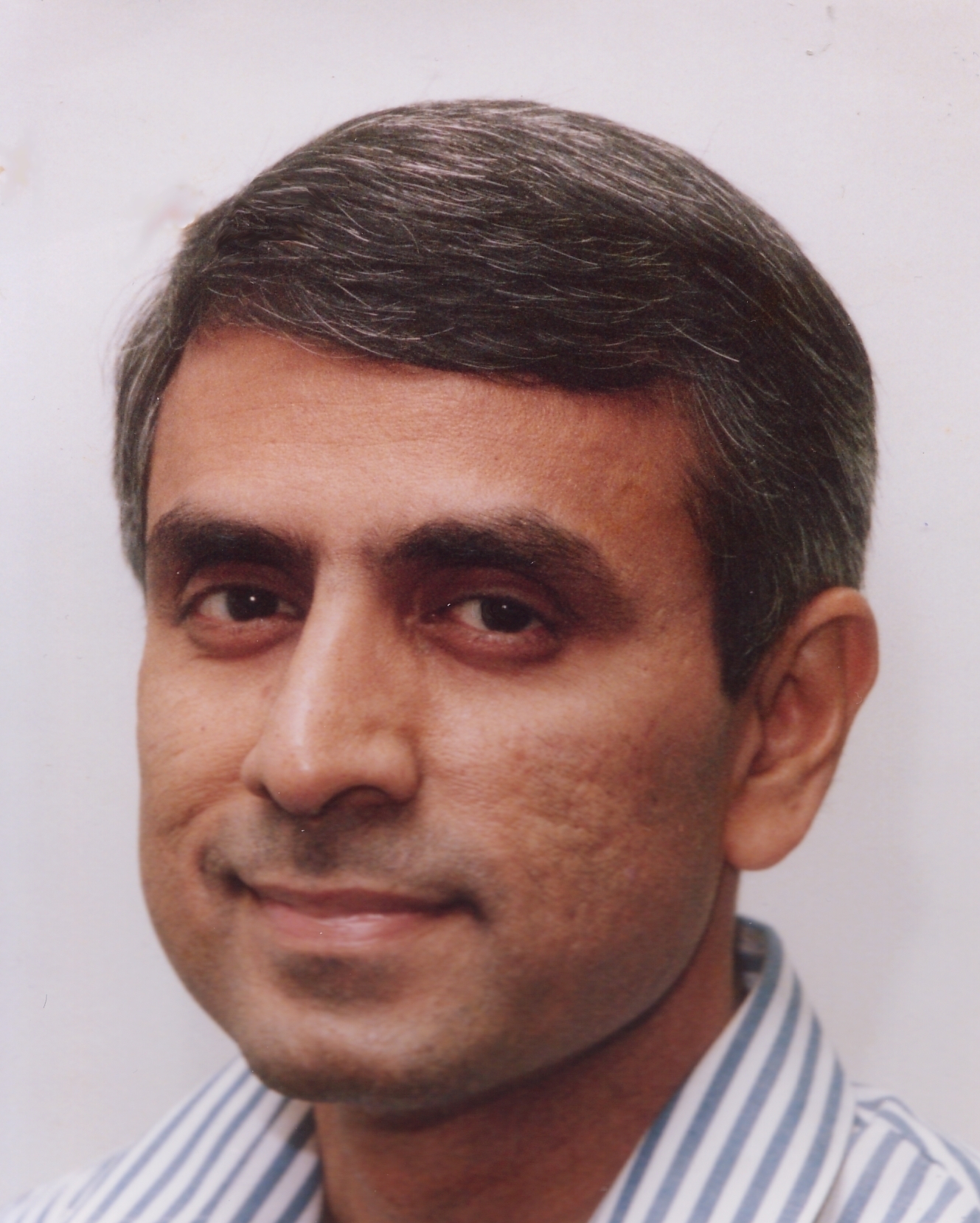}}]{Sridha Sridharan} has obtained an MSc (Communication Engineering) degree from the University of Manchester, UK, and a PhD degree from the University of New South Wales, Australia. He is currently with the Queensland University of Technology (QUT) where he is a Professor in the School of Electrical Engineering and  Robotics. He has published over 600 papers consisting of publications in journals and in refereed international conferences in the areas of Image and Speech technologies during the period 1990-2023.  During this period he has also graduated 85  PhD students in the areas of Image and Speech technologies. Prof Sridharan has also received a number of research grants from various funding bodies including the Commonwealth competitive funding schemes such as the Australian Research Council (ARC) and the National Security Science and Technology (NSST) unit. Several of his research outcomes have been commercialised.
\end{IEEEbiography}

\vfill

\end{document}